# Democratising AI: Multiple Meanings, Goals, and Methods

Elizabeth Seger[1]
Centre for the Governance of AI
Oxford, UK
elizabeth.seger@governance.ai

Aviv Ovadya
Harvard Berkman Klein Centre
Cambridge, MA
aviv@aviv.me

Ben Garfinkel
Centre for the Governance of AI
Oxford, UK
ben.garfinkel@governance.ai

Divya Siddarth
Collective Intelligence Project
Oxford, UK
divya@cip.org

Allan Dafoe
Google DeepMind
London, UK
allandafoe@deepmind.com

## ABSTRACT

Numerous parties are calling for "the democratisation of AI", but the phrase is used to refer to a variety of goals, the pursuit of which sometimes conflict. This paper identifies four kinds of "AI democratisation" that are commonly discussed: (1) the democratisation of AI use, (2) the democratisation of AI development, (3) the democratisation of AI profits, and (4) the democratisation of AI governance. Numerous goals and methods of achieving each form of democratisation are discussed. The main takeaway from this paper is that AI democratisation is a multifarious and sometimes conflicting concept that should not be conflated with improving AI accessibility. If we want to move beyond ambiguous commitments to "democratising AI", to productive discussions of concrete policies and trade-offs, then we need to recognise the principal role of the democratisation of AI governance in navigating tradeoffs and risks across decisions around use, development, and profits.

## 1 Introduction

Over the last couple years, discussion of "AI democratisation" has surged. AI companies around the world—such as Stability AI [1][1], Meta [2], Microsoft [3], and Hugging Face [4]—are talking about their commitment to democratising AI, but it's not always clear what they mean. The term "AI democratisation" seems to be employed in a variety of ways, causing commentators to speak past one another when discussing the goals, methodologies, risks, and benefits of AI democratisation efforts. This paper aims to provide a foundation for more productive conversations about democratising AI that move beyond ambiguous commitments.

Sections 2 through 5 describe four different notions of AI democratisation commonly used by AI labs—democratisation of AI use, democratisation of AI development, democratisation of AI profits, and democratisation of AI governance. We focus primarily on how the term "AI democratisation" is used by AI labs because of the impact they have on the rate of AI advances, and the influence they currently wield over the use, development, profit, and governance of AI. If labs are claiming commitments to AI democratisation, then it's important to clarify what those commitments mean and how they might be fulfilled.

Each section is divided into two subsections. The first subsection (x.1) discusses various goals the particular form of democratisation is proposed to achieve and notes conflicts with the goals of other forms of democratisation where they arise. The second subsection (x.2) describes various proposed methods for facilitating the form of democratisation.

Although the four concepts of democratisation we discuss often complement each other, it is important to note that they sometimes conflict. For instance, if the public prefers for access to certain kinds of AI systems to be restricted, then the "democratisation of AI governance" may require access restrictions to be put in place—but enacting these restrictions may hinder the "democratisation of AI development" for which some degree of AI model accessibility is key.

---

[1] **Author Contributions:** ES conceived of most of the presented ideas and frameworks, and drafted most of the manuscript with iterative feedback and support from AO, BG, and AD. AO conceptualised and drafted most of Section 5.2. DS conceptualised Section 4 and provided editing support.

Section 6 then concludes, driving home the main observation of the paper; though the term "democratisation" can seem to imply otherwise, AI democratisation is not inherently good. The first three forms of democratisation (democratisation of use, development, and profits) are about improving accessibility to AI or AI derived profits which can yield both beneficial and harmful consequences. The desirability of AI democratisation therefore can not be assumed, but rather is derived from alignment with the interests and values of those who will be impacted.

## 2 Democratisation of AI Use

When people speak about democratising some technology, they often refer to democratising its use—that is, making it easier for a wide range of people to use the technology. For example, in the early 2010's the "democratisation of 3D printers" referred to how 3D printers were becoming much more easily acquired, built, and operated by the general public [5].

The same meaning has been applied to the democratisation of AI. Stability AI, for instance, has been a vocal champion of AI democratisation. The company proudly describes its main product, the image generation model Stable Diffusion, as "a text-to-image model that will empower billions of people to create stunning art within seconds" [6]. Microsoft similarly claims to be undertaking an ambitious effort "to democratize Artificial Intelligence (AI), to take it from the ivory towers and make it accessible to all." A salient part of its plan is "to infuse every application that we interact with, on any device, at any point in time, with intelligence" [3].

### 2.1 Goals

*2.2.1 Distributing Benefits of Use*

The most commonly articulated goal of democratising AI use, and that communicated above, is to distribute benefits of AI use for many people to enjoy. Benefits include entertainment value (e.g. generating poems with ChatGPT), health and well-being applications, productivity improvement, and other utility functions (writing code, analysing data, creating art). Many of these benefits might, in turn, be translated into financial gains for those who effectively integrate the AI tools into their workstreams. That said, access to these benefits is often financially gated; AI companies that aim to democratise use of their products generally offer their more powerful tools and support services at a price, by purchase or paid subscription.

Ensuring widespread accessibility to some kinds of AI tools may serve also as an important equalising opportunity, providing people with access to services (e.g. medical advice) they would otherwise find difficult to acquire due to financial, geographical, or transportation restrictions. Similarly, tools like ChatGPT that can be used to draft formal statements and letters may help people to formally vocalise complaints who might otherwise have felt disenfranchised by poor education or limited grasp of the relevant language.

It is important to recognize, however, that for some AI applications the benefits of making the technology available for anyone to use can be relatively minor while the risks are significant. For example, the circle of individuals who would greatly benefit from access to an AI drug discovery tool is relatively small (mainly pharmaceutical researchers), however these tools can be easily repurposed to discover new toxins that might be used as chemical weapons [7]. This is an instance in which unfettered democratisation of AI use—making an AI tool accessible to all—may not always be desirable.

That said, an AI tool need not be widely accessible to all for the benefits to be widely distributed. A designated user could employ a high-risk AI system for the benefit of the community. In this way a drug discovery system could be used in a controlled, limited-access setting, while resulting pharmaceuticals are "democratised" in the sense that they are made accessible to anyone in need.

*2.2.2 Receiving Feedback for Better and Safer AI*

Another reason given for disseminating AI tools widely is so that developers can gather information about how their products are being used (or misused) in a wider variety of contexts than they would have been able to test, let alone imagine, internally [8]. In turn, that feedback informs improvements that enhance model performance and help guard against any new misuse applications that have emerged.

Importantly, where there are concerns about potential misuse, there is an option to cautiously democratise AI use via a staged release of the product [9]. Incrementally larger and more powerful versions of the model are released allowing time between each stage to evaluate how the AI application is being used and to conduct risk benefit analyses of releasing yet a more powerful version of the model. Feedback and staged release may also help provide time and notice for societies to adapt to the new capabilities and harden vulnerable systems, processes, and institutions.

Unfortunately, where the risk and consequences of misuse are expected to be severe, responsible AI deployment may require that access restrictions be placed on certain AI capabilities.

Such restrictions limit the democratisation of AI use, but they are not necessarily a blow to AI democratisation more generally. As will be discussed in Section 5, AI democratisation can also refer to the democratisation of AI governance, which is about introducing democratic processes into decision-making about how AI is used, developed, shared and otherwise regulated. Indeed, one might say that the democratisation of AI use—making AI accessible to be used by everyone—is but one possible outcome of democratising AI governance. It is the outcome if the demos choose that distributing use is desirable. Another possible outcome may be to designate (perhaps through licensing) specific actors to use or study high-risk AI systems for the public's benefit.

## 2.2 Methods

Overall, efforts to democratise AI use—to make AI capabilities more widely accessible—may involve reducing the costs of acquiring and running AI tools (again, this may be done via staged release if there are concerns about misuse), providing accessible services to help users integrate AI models into their work streams (e.g. consulting services), and developing intuitive interfaces to facilitate human-AI interaction without the need for extensive training or technical knowhow. In some regions, democratising AI use may also require improvement to more fundamental infrastructure like internet access [10].

## 3 Democratisation of AI Development

When the AI community talks about democratising AI, they rarely limit their focus to the democratisation of AI use. Much of the excitement is about democratising AI development—that is, helping a wider range of people contribute to AI design and development processes.

### 3.1 Goals

*3.1.1 Accelerate AI innovation and Progress*

A common theme is that tapping into a global community of potential contributors will accelerate innovation and progress in AI research and development. As Microsoft CTO Kevin Scott explains, "we at tech companies, in Silicon Valley startups, can't possibly imagine all of the interesting and valuable things that can be done with [AI], and that we need not just tens or hundreds of thousands of people who work at these companies to be building things but we need hundreds of millions or billions of people able to harness the power of these machines" [11].[2]

It should not be assumed, however, that accelerating AI progress is always desirable. As AI research progresses and AI capabilities improve, we should expect the potential consequence of harms, misuse, or misalignment to also become more severe [12], [13]. The implementation of necessary policy and interventions to ensure safe and responsible AI development going forward may struggle to keep up with unbridled progress, and so there may be a case for exercising some restraint [14].

*3.1.2 Cater to Diverse Interests and Needs*

Calls for the democratisation of AI development also respond to concern that a small number of leading AI labs monopolise control over AI development and that those labs employ a narrow demographic of developers. The worry is that the AI products, which are deployed globally, consequently perform disparately for users of different ethnic, geographic, cultural, professional, and financial backgrounds [15]–[18].

Enabling more people to participate in AI design and development processes may help facilitate the development of AI applications that cater to more diverse interests and needs [19]. This is one reason Stability AI offers for its decision to open-source Stable Diffusion—meaning that the company allows anyone to download, modify, or build on the Stable Diffusion model on their own computer so long as they agree to the terms of use. CEO Emad Mostaque advocates that "everyone needs to build [AI] technology for themselves…. It's something that we want to enable because nobody knows what's best [for example] for people in Vietnam besides the Vietnamese" [1].[3] The company motto reads "AI by the people, for the people" [20].

But again, it should not be assumed that the diffusion of AI development is universally desirable. Open-source sharing in particular may enable more numerous and diverse contributions, but it also opens a door for malicious use and model modification, and controls are difficult to enforce [21].

*3.1.3 External Evaluation*

Third, many argue that involving more people (e.g. academics, individual developers, smaller labs) in AI development processes provides a critical external evaluation and auditing mechanism. By making models accessible for more people to study, AI labs might distribute auditing duties to a larger and more diverse group of developers than a lab would be able to employ internally. This assumes that more eyes on a model will reveal more flaws leading to safer and more well-aligned technology [8].

---

[2] Quote occurs at 22:30
[3] Quote occurs at 13:00

## 3.2 Methods

A variety of activities can help enable productive participation in AI design and development processes. Some strategies provide access to AI models and resources to facilitate AI community engagement—e.g. model sharing (3.2.1), providing compute access (3.2.2), project support and coordination (3.2.3). Other strategies help to expand the community of people capable of contributing to AI development processes—e.g. via educational & upskilling opportunities (3.2.4) or through the provision of assistive tools (3.2.5).

A key takeaway from this section is that there is much more to AI democratisation—even to the democratisation of AI development specifically—than model dissemination.

### 3.2.1 Model Sharing

Model sharing involves providing access to AI models including code, model weights and the ability to query, modify, study, or otherwise examine the model.

While model sharing is needed to enable external study and auditing of AI models, it also increases the opportunity for model misuse by malicious actors. In some contexts, for higher risk capabilities, it may therefore be wise to limit model accessibility [22].

That said, model sharing is not an all or nothing activity [23]. Rather, model sharing options range from full open-source sharing (all aspects of the system are downloadable for the public) to fully closed (only a select group of developers may even know the model exists) [21], [22], [24]. In the middle there are options for gated access, hosted model access, cloud-based or API access, and downloadable access with some model components withheld. In some of these middle options some degree of advantage from external study and auditing may be maintained while risk of misuse might be reduced.

These options should not be interpreted naively with respect to their stated intent, but with realism about their likely impacts. Google Research published a paper on a technique for style cloning in generative art, and chose to not release any code, citing the potential "societal impact" risk that "malicious parties might try to use such images to mislead viewers" [25]. However, in a mere 11 days one person was able to reproduce the technique to run on Stable Diffusion which they then chose to open-source [26]. Similarly, Meta chose to restrict access to the weights of its large language model LLaMa to academic researchers and others on a case by case basis, "to maintain integrity and prevent misuse" [27]. However, a week later the weights (predictably) were leaked and are now available publicly on a torrent [28]. In both these cases, realism is needed in assessing the likely impact of a nominally more restrictive model sharing policy.

### 3.2.2 Improving Compute Access (and other technical infrastructure)

Large AI models require significant compute power to run. Accordingly, democratising development may also require improvements to compute access. Developers might, for instance, offer cloud computing services or issue grants for computer cluster access to facilitate smaller and less well-resourced groups in working with more powerful models [21]. Alternatively, developers might explore options for providing smaller model versions that require less compute to run. For example, Emad Mostaque describes Stable Diffusion as "a breakthrough in speed and quality [...] that can be run on consumer GPU's" [6].

Note, however, that restrictions on compute can also be leveraged to help minimise misuse of powerful AI by limiting the ability of prospective malicious actors to build or modify large models [29], [30]. Therefore, like decisions to open-source AI models, decisions to provide significant compute resources should involve adequate risk benefit analysis.

Other tech infrastructure that limits participation in AI development in a similar way to compute access include network accessibility (i.e. access to high bandwidth, low latency internet), access to data storage facilities, access to high quality ethically sourced data, and cyber security infrastructure. These all pose significant barriers to participation in AI development in resource constrained countries, barriers which might be lessened through infrastructure investment and/or remote access [10].

### 3.2.3 Project Support and Coordination

Democratising AI development is not just about providing resources and assuming that people will come. Effective input elicitation often benefits from dedicated project coordination and support.

For example, the BigScience project was a collaborative effort coordinated by the AI startup Hugging Face—another organisation dedicated to "democratising

AI"—and funded by the French government to develop the large language model (LLM) BLOOM [31]. BLOOM was developed over the course of a year by a global coalition of over 1000 volunteer AI developers yielding an LLM functional in 46 languages. Similar collective efforts in other domains may also benefit from funding or other resources to support coordination.

*3.2.4 Educational and Upskilling Opportunities*

Democratisation of AI development can also be furthered by expanding the community of people capable of making contributions to AI design and development processes.

One option toward this end is for governments and large developer labs to invest in making educational and upskilling opportunities more widely available, especially for demographics traditionally underrepresented in AI developer communities. Investment in computer science and machine learning educational resources is, for instance, seen as an essential step for establishing AI talent pipelines and narrowing the 'AI divide' between the Global North and South [10].

*3.2.5 Assistive Tools*

Another option for expanding the community of prospective contributors is to lower barriers to participation in AI development activities by making it easier for people with minimal programming experience and little familiarity with machine learning to partake.

This might be done through the provision of tools that enable those with less experience and expertise to create and implement their own machine learning applications. For example, Microsoft [32], Google [33], H2O [34] and Amazon [35] have developed "no-code" tools that allow people to build models that are personalised to their own needs without prior coding or machine learning experience. In a similar vein, GitHub Copilot (powered by OpenAI Codex) is a generative AI system that can be used by less experienced developers to help write code [36].[4]

## 4 Democratisation of AI Profits

A third sense of "AI democratisation" refers to democratising AI profits—which is about facilitating the broad and equitable distribution of value accrued to organisations that build and control advanced AI capabilities.

The notion is nicely articulated by Microsoft's CTO Kevin Scott: "I think we should have objectives around real democratisation of the technology. If the bulk of the value that gets created from AI accrues to a handful of companies in the West Coast of the United States, that is a failure" [11].[5] Though DeepMind does not employ "AI democratisation" terminology, CEO Demis Hassabis expresses a similar sentiment. As reported by TIME, Hassabis believes the wealth generated by advanced AI technologies should be redistributed. "I think we need to make sure that the benefits accrue to as many people as possible—to all of humanity, ideally" [38].

### 4.1 Goals

The goal is rather straightforward: equitably distribute profits generated by AI to ensure wealth and advantages conferred by AI improve human well-being across the board. A few sub-aims are: to avoid widening a socioeconomic divide between AI leading and lagging nations [10]; to ease the financial burden of job loss to automation; to smooth economic transition in case of rapid growth of the AI industry; and, when AI labs are able to voluntarily participate, to provide mechanisms for labs to powerfully demonstrate their commitment to pursuing advanced AI for the common good [39], [40]. Finally, profit democratisation acknowledges through compensation the human labour and creativity that underpins AI capabilities. Generative AI, in particular, unlocks economic value in training data that has been produced through centuries of human effort.

### 4.2 Methods

There are a variety of mechanisms by which AI profits might be more widely distributed or "democratised". Profits might be redistributed, for instance, via philanthropic giving, though philanthropy can be an inconsistent mechanism of wealth redistribution and, if not well-managed, may worsen inequalities and injustices [41].

Another option is for taxation and profit redistribution to be managed directly by the state [42]. For example, the provision of Universal Basic Income (UBI) has been suggested as a wealth distribution mechanism to help compensate for job loss to automation associated with more advanced AI capabilities [39], [43].

There is concern, however, that taxation methods may be insufficient given the potential of monopolised windfall profits to major AI labs. Accordingly, the proposed "Windfall Clause" offers a third, middle-ground approach [40]. AI firms that voluntarily adopt the Windfall Clause would be bindingly obliged to donate a meaningful portion of their profits when the firm's profits for the year exceed "a substantial fraction of the world's total economic output" (e.g. at least 1%). Those donations would then go to a "Distributor" charged with finding and funding effective welfare-maximising projects. Distributors might offer grants to philanthropic organisations, invest directly in infrastructure building projects, or direct funds to state governments for further distribution.

---

[4] Though there is concern that users are overconfident and write less secure code when relying on AI assistants like Copilot [37].

[5] Quote occurs at 47:30

Finally, there is a question of if and how individual content creators can be compensated when their creative outputs (art, music, code, etc.) are used to train generative AI models [44]. One option is through the creation of licensed data sets [45]; content creators are compensated for permitting their content to be included in a catered data set that AI developers can then use to train and fine-tune their models without risk of copyright infringement. However, there is still an open question as to if and how further compensation should be provided as generative AI continues to produce value after it is trained. Here is perhaps where the more general profit redistribution schemes described above play an important role.

# 5 Democratisation of AI Governance

Finally, some discussions about AI democratisation refer to democratising AI governance. AI governance decisions often involve balancing AI-related risks and benefits to determine if, how, and by whom AI should be used, developed, and shared. The democratisation of AI governance is about distributing influence over these decisions to a wider community of stakeholders and impacted populations. OpenAI CEO Sam Altman has expressed such a sentiment, writing, "We want the benefits of, access to, and governance of AGI to be widely and fairly shared" [46].

## 5.1 Goals

The overarching goal of the democratisation of AI governance is to ensure that decisions around questions such as AI usage, development, and profits reflect the will and preferences of the people being impacted [47]. In this sense, democratisation of AI governance arguably supersedes the previously discussed notions of democratisation. Decisions to democratise use, development, and profits derive their acceptability and desirability from the acceptance and desire of those who will be impacted.

Democratic processes such as referenda, citizen assemblies, and public hearings facilitate the representation of diverse and often conflicting beliefs, opinions, and values into decisions about how people and their actions are governed.

Importantly, the desired result is not necessarily agreement among constituents that the best decision was made, but legitimacy—a state of acceptance that the decision-making process was fair and well-considered.

### 5.1.1 Reducing Unilateral Decision-making

Motivation for democratising AI governance often stems from concern that individual tech companies hold unchecked control over the future of a transformative technology and too much freedom to decide for themselves what constitutes acceptable tradeoffs between risks and benefits of choices around AI use, development, and distribution of profits. It is a worry exacerbated by concern that fierce competition between leading AI labs incentivises reckless decision-making [48].

A single actor in control of a powerful technology or resource can cause significant harm with an ill-considered decision. Consider, for example, the avoidable 2010 Deepwater Horizon oil spill in the Gulf of Mexico. It is one of the greatest environmental disasters in history and largely attributed to a series of cost cutting decisions made by BP including failure to implement proper risk control measures [49]. It is reasonable to assume that in the face of fierce competition and massive financial incentive that AI developers are also liable to make rash decisions about model development and release, the potential negative repercussions of which will only grow as AI capabilities improve [12], [13]. Therefore, it is perhaps unwise, as Stability AI CEO Emad Mostaque puts it, to have "a centralised, unelected entity controlling the most powerful technology in the world" [50]. Introducing democratic processes to enable checks and balances or collective decision-making around AI development, use, and release can potentially guard against ill-considered and potentially detrimental moves. Though Mostaque was justifying Stability AI's decision to open-source its models as a method of disseminating control over AI, not commenting on how or by whom such a high-stakes decision should be made.

### 5.1.2 Justice and Fairness

Another commonly articulated goal is to ensure the benefits and burdens of AI development and deployment are distributed justly and fairly.

It is widely documented that AI systems can replicate or even amplify racial and societal injustices [51], for example, through algorithmic bias in hiring [15], facial recognition [16], loan appraisal [18], and recidivism prediction applications [17]. Some AI misuse cases, such as voice cloning-based phishing, may also have a disproportionate effect on some populations over others [52].

Overall, facilitating the participation or representation of a wide array of stakeholders is seen as a crucial step towards mitigating AI associated injustices [53], [54]. It is to work towards a future for AI in which no communities are disproportionately harmed by development and use activities, and in which no communities are unfairly overlooked as possible beneficiaries of AI capabilities.

### 5.1.3 Navigating Complex Normative Challenges

The implementation of AI systems in public and private applications raises a variety of normative questions. Some are readily agreed upon such as the high-level assertion that human fatality should be avoided. But there are also many "hard normative questions" to which responses will likely differ depending on culture, context and other value priorities [55]. These challenges include, for example, establishing acceptable risk thresholds, interpreting high-level terms like the above mentioned "justice" and "fairness", and determining what values should underpin value-aligned AI [47]. Democratic discourse among diverse stakeholders may help distil areas of agreement or areas in which consensus forming practices are likely to be productive [47]. Inversely, and of equal importance, they might identify cases in which finer-grained details of interpretation and implementation can be determined at a context-specific, local level [55].

## 5.2 Methods

Even though it is already a subcategory of AI democratisation, the democratisation of AI governance is itself a broad and multifaceted concept, some forms of which may be more relevant or useful than others depending on the context [56].

One might speak, for instance, about the introduction of democratic processes to high-level AI policy formation at the national or international governance level or about more fine-grained AI design or deployment decisions made within individual labs [57]. "Democratic processes" can also refer to a variety of methods for eliciting citizen participation, ensuring substantive representation of stakeholder viewpoints, facilitating well-informed deliberation, holding fair and open election processes, or instituting constitutional protections for individuals and minorities [58].

In what follows, we briefly describe a variety of strategies that have been proposed to underpin democratically legitimate decision making about AI.

### 5.2.1 Harnessing Existing Democratic Structures

Democratic societies already have many tools and infrastructures in place to facilitate democratically legitimate decision-making about a variety of topics through e.g. legislation and regulation or multilateral standards. Harnessing and modifying effective structures already in place avoids redundancy and reinventing the wheel [57]. It has been proposed, for instance, that with some modification we might make use of procedures laid out by the European Union's standard-setting organisations (SSOs) to establish context-sensitive standards for safe and responsible AI [55]. We might also structure a new AI governing body after the United States' FDA (Federal Food and Drug Administration) [59].

While such efforts require minimal new infrastructure, they can, however, get bogged down in existing political quagmires, and are only applicable for decisions that remain within the borders of democratic societies.

### 5.2.2 Multistakeholder Bodies

Given the global impacts of many AI advances, there has been significant interest in the use of more inclusive processes for input and decision-making around AI. One option is through the formation of multistakeholder bodies to convene diverse, international perspectives for the purpose of navigating complex AI governance challenges. For example, the Partnership on AI (PAI)—a global coalition of academic, civil society, industry, and media organisations—has orchestrated initial non-binding agreements on generative AI across some relevant organisations [60]. There have also been proposals for smaller multistakeholder bodies to form the basis of ethical review boards for high risk AI application and model release decisions [24], [61].

### 5.2.3 Participatory Processes

Another strategy is to employ modern participatory processes to gather input from diverse populations to guide AI governance decisions. Orchestrating large scale public participation can be cumbersome and costly, so much promising work focuses on exploring technical solutions such as deliberative tools and digital platforms [62] and generative voting applications [63] to improve the practicability and accessibility of participatory AI governance.

A disadvantage with a participatory approach to governance, however, is that those involved are generally self-selected as stakeholders, community members, or participants, and their outputs may thus only have a weak claim to democratic legitimacy.

*5.2.4 Representative Deliberation*

A strategy for addressing the challenges of both transnational AI impacts and self-selection is the use of representative deliberation [64], [65], building on heavily researched approaches to deliberative democracy [66]. Representative deliberation involves putting AI governance questions to a representative microcosm of the population of an impacted region, or even the global population (selected by sortition, i.e. stratified sampling) thus granting democratic legitimacy.[6] As is common practice with citizen assemblies, the representative groups are provided access to experts and stakeholders to help inform their deliberations on more technical topics such as AI governance.

Representative deliberation is increasingly lauded by both governments [67] and multilateral bodies [68] as a valuable modern approach to democracy generally, and which has found footholds even in authoritarian and global contexts under the less threatening frame of deliberative governance [66], [69].

Companies developing AI systems that want to 'democratise their governance' can also delegate such decisions to representative deliberations, and often will have the incentive to do so in the face of competing stakeholder pressures [62], [65]. Meta, for example, has quietly run a set of national and transnational pilots [70] to navigate their 'complex normative challenges' and have since scaled up to a near-global deliberative process [71]. Twitter had also planned to pilot such processes before its acquisition [72].

Lighter weight variants of representative deliberations that build on the aforementioned modern participatory practices but in a representative fashion (e.g. using sortition) might also be used to provide a level of democratic legitimacy for less complex AI governance questions [62], [63].

# 6 Conclusion

This paper has outlined four different notions of "AI democratisation"—the democratisation of use, the democratisation of AI development, the democratisation of AI profits, and the democratisation of AI governance—and discussed numerous goals and methods of achieving each.

For the first three forms of democratisation, "democratisation" is used almost synonymously with "increasing accessibility". The democratisation of AI use and the democratisation of AI development are about making AI systems accessible for everyone to use or to contribute to their development, and the democratisation of AI profits is about distributing access to profits accrued through AI development and control. The democratisation of AI governance, however, is about balancing these questions of accessibility with other societal needs and values.

Sometimes decisions to democratise AI use, development, and profits will align with societal preferences (ideally determined through democratic processes), and sometimes those preferences will involve restrictions on access. Such is the case, for instance, with legal restrictions on certain medications, treaty restrictions on nuclear weapons, and regulation of labs containing potential hazards. The same may be true of decisions to restrict access to AI models for development or use purposes if risks of open model access are felt to outweigh the benefits.

We should therefore be wary of using the term "democratisation" too loosely or, as is often the case, as a stand-in for "all things good". The democratisation of AI use, AI development, and even AI derived profits are not inherently good.[7] Their value is derived from alignment with interests and values of those who will be impacted. As such, where the democratically aligned decision would be to limit accessibility, the democratisation of AI governance takes precedence over the others as the source from which the moral and political value of the "democratisation" terminology is derived.

Perhaps the proper response to this paper, then, is to conclude that "AI Democratisation" is a (mostly) unfortunate term. As it is most commonly used within the AI community it refers to facilitating widespread AI use and development. However, invoking the term "democratisation" tells another story. It holds the hidden assumption that the decision to distribute or make accessible is what a democratic governance process would select. In other words, AI democratisation ultimately refers to the democratisation of AI governance. If by "AI democratisation" all a speaker means is "make available to everyone", then we would suggest less normatively loaded language (something like "broad accessibility") be used.

## ACKNOWLEDGEMENTS

*We would like to thank Toby Shevlane, Sammy McKinny, Leonie Koessler, Ben Harack, Markus Anderljung, Lennart Heim, Noemi Dreksler, Anton Korinek, Guive Assadi, and*

---

[6] Multistakeholder and participatory inputs can integrate with representative deliberations, but the ultimate recommendations are decided on by the representative microcosm, thus granting a stronger claim to democratic legitimacy than multistakeholder or participatory approaches alone.

[7] Improperly managed profit redistribution schemes may worsen inequalities and injustices [41].